\documentclass[11pt,a4paper]{article}

\usepackage[hyperref]{acl2018} 

\usepackage[utf8]{inputenc}  
\usepackage{times}

\usepackage{latexsym}

\usepackage{url}

\aclfinalcopy 
\def\aclpaperid{36} 

\usepackage{amsmath}
\usepackage{graphicx}
\usepackage{adjustbox}
\usepackage{subcaption}
\usepackage{amssymb}
\usepackage{multirow}
\usepackage{boxedminipage}
\usepackage{rotating}

\usepackage{mathtools}
\usepackage{xspace}
\usepackage{afterpage}
\usepackage{pdflscape}
\usepackage{array}
\usepackage{arydshln}
\usepackage{float}
\usepackage{dcolumn}

\usepackage{xcolor}
\definecolor{darkblue}{rgb}{0, 0, 0.5}
\usepackage[normalem]{ulem}
\definecolor{darkgreen}{rgb}{0.0, 0.5, 0.0}

\def\VRdel#1{\bgroup\markoverwith{\textcolor{magenta}{\rule[0.5ex]{2pt}{1pt}}}\ULon{#1}}

\def\ODdel#1{\bgroup\markoverwith{\textcolor{darkgreen}{\rule[0.5ex]{2pt}{1pt}}}\ULon{#1}}

\newcommand\tgen{\textsc{TGen}\xspace}
\newcommand\slug{\textsc{Slug}\xspace}
\newcommand\slugalt{\textsc{Slug-alt}\xspace}
\newcommand\tntnlgi{\textsc{TNT1}\xspace}
\newcommand\tntnlgii{\textsc{TNT2}\xspace}
\newcommand\zhawi{\textsc{ZHAW1}\xspace}
\newcommand\zhawii{\textsc{ZHAW2}\xspace}
\newcommand\adapt{\textsc{Adapt}\xspace}
\newcommand\dangnt{\textsc{DANGNT}\xspace}
\newcommand\forgei{\textsc{FORGe1}\xspace}
\newcommand\forgeiii{\textsc{FORGe3}\xspace}
\newcommand\gong{\textsc{Gong}\xspace}
\newcommand\harv{\textsc{Harv}\xspace}
\newcommand\nle{\textsc{NLE}\xspace}
\newcommand\sheffi{\textsc{Sheff1}\xspace}
\newcommand\sheffii{\textsc{Sheff2}\xspace}
\newcommand\chen{\textsc{Chen}\xspace}
\newcommand\thomsoni{\textsc{TR1}\xspace}
\newcommand\thomsonii{\textsc{TR2}\xspace}
\newcommand\tuda{\textsc{TUDA}\xspace}
\newcommand\zhang{\textsc{Zhang}\xspace}

\definecolor{seqtoseq}{rgb}{0.941, 0.062, 0.207}
\definecolor{datadriven}{rgb}{0.960, 0.509, 0.188}
\definecolor{rules}{rgb}{0.156, 0.784, 0.231}
\definecolor{templates}{rgb}{0, 0.4, 0.619}
\newcommand{\symbseq}{$^\heartsuit$}
\newcommand{\symbdd}{$^\diamondsuit$}
\newcommand{\symbrule}{$^\clubsuit$}
\newcommand{\symbtempl}{$^\spadesuit$}

\newcommand\Ctgen{\textcolor{seqtoseq}{\symbseq\bf \tgen}}
\newcommand\Cslug{\textcolor{seqtoseq}{\symbseq\bf \slug}}
\newcommand\Cslugalt{\textcolor{seqtoseq}{\symbseq\bf \slugalt}}
\newcommand\Ctntnlgi{\textcolor{seqtoseq}{\symbseq\bf \tntnlgi}}
\newcommand\Ctntnlgii{\textcolor{seqtoseq}{\symbseq\bf \tntnlgii}}
\newcommand\Czhawi{\textcolor{datadriven}{\symbdd\bf \zhawi}}
\newcommand\Czhawii{\textcolor{datadriven}{\symbdd\bf \zhawii}}
\newcommand\Cadapt{\textcolor{seqtoseq}{\symbseq\bf \adapt}}
\newcommand\Cdangnt{\textcolor{rules}{\symbrule\bf \dangnt}}
\newcommand\Cforgei{\textcolor{rules}{\symbrule\bf \forgei}}
\newcommand\Cforgeiii{\textcolor{templates}{\symbtempl\bf \forgeiii}}
\newcommand\Cgong{\textcolor{seqtoseq}{\symbseq\bf \gong}}
\newcommand\Charv{\textcolor{seqtoseq}{\symbseq\bf \harv}}
\newcommand\Cnle{\textcolor{seqtoseq}{\symbseq\bf \nle}}
\newcommand\Csheffi{\textcolor{datadriven}{\symbdd\bf \sheffi}}
\newcommand\Csheffii{\textcolor{seqtoseq}{\symbseq\bf \sheffii}}
\newcommand\Cchen{\textcolor{seqtoseq}{\symbseq\bf \chen}}
\newcommand\Cthomsoni{\textcolor{seqtoseq}{\symbseq\bf \thomsoni}}
\newcommand\Cthomsonii{\textcolor{templates}{\symbtempl\bf \thomsonii}}
\newcommand\Ctuda{\textcolor{templates}{\symbtempl\bf \tuda}}
\newcommand\Czhang{\textcolor{seqtoseq}{\symbseq\bf \zhang}}

\newcolumntype{F}{D{.}{.}{2,3}}

\title{Findings of the E2E NLG Challenge}  

\author{Ondřej Dušek, Jekaterina Novikova \and Verena Rieser \\
  The Interaction Lab, School of Mathematical and Computer Sciences \\
  Heriot-Watt University \\
  Edinburgh, Scotland, UK \\
  {\tt \{o.dusek, j.novikova, v.t.rieser\}@hw.ac.uk} \\
}

\date{}

\begin{document}
\maketitle
\begin{abstract}%
This paper summarises the experimental setup and results of the first shared task on end-to-end (E2E) natural language generation (NLG) in spoken dialogue systems. Recent end-to-end generation systems are promising since they reduce the need for data annotation. However, they are currently limited to small, delexicalised datasets. 
The E2E NLG shared task aims to assess whether these novel approaches can generate better-quality output by learning from a dataset containing higher lexical richness, syntactic complexity and diverse discourse phenomena. 
We compare 62 systems submitted by 17 institutions, covering a wide range of approaches, including machine learning architectures 
-- with the majority implementing sequence-to-sequence models (seq2seq) --
as well as systems based on grammatical rules and templates.
\end{abstract}

\section{Introduction}

This paper summarises the first shared task on 
 end-to-end (E2E) natural language generation (NLG) in spoken dialogue systems (SDSs). 
Shared tasks have become an established way of pushing research boundaries in the field of natural language processing, with NLG benchmarking tasks running since 2007 \citep{belz:GRE2007}. This task is novel in that it poses new challenges for 
recent end-to-end, data-driven NLG systems for SDSs which 
jointly learn sentence planning and surface realisation and do not require costly semantic alignment between meaning representations (MRs) and the corresponding natural language reference texts, e.g.\ \cite{jurcicek:2015:ACL,wen:emnlp2015,Mei:NAACL2016,Wen:NAACL16,SharmaHSSB16,Dusek:ACL16,vlachos:coling2016}.%
\footnote{Note that as opposed to the ``classical'' definition of NLG \cite{reiter_building_2000,gatt_survey_2018}, generation for dialogue systems does not involve content selection and its sentence planning stage may be less complex.}
So far, end-to-end approaches to NLG are limited to small, delexicalised datasets, e.g.\ BAGEL \citep{mairesse:acl2010}, SF Hotels/\hspace{0mm}Restaurants \citep{wen:emnlp2015}, or RoboCup \citep{chen2008learning}, whereas the E2E shared task is
based on a new crowdsourced dataset of 50k instances in the restaurant domain, which is about 10 times larger and also more complex than previous datasets.
For the shared challenge, we received 62 system submissions by 17 institutions from 11 countries, with about $1/3$ of these submissions coming from industry. 
We assess the submitted systems by comparing them to a challenging baseline using automatic as well as human evaluation.
We consider this level of participation an unexpected success, which underlines the timeliness of this task.\footnote{In comparison, the well established Conference in Machine Translation WMT’17 (running since 2006) received submissions from 31 institutions to a total of 8 tasks \citep{bojar2017findings}.}
While there are previous studies comparing a limited number of end-to-end NLG approaches \cite{Novikova:EMNLP2017,Wiseman:EMNLP17,WebNLG},
this is the first research to evaluate novel end-to-end generation at scale and using human assessment. 


\section{The E2E NLG dataset}\label{sec:dataset}

\subsection{Data Collection Procedure}\label{sec:collection}

In order to maximise the chances for data-driven end-to-end systems to produce high quality output, we aim to provide training data in high quality and large quantity. 
To collect data in large enough quantity, we use crowdsourcing with automatic quality checks. 
We use MRs consisting of an unordered set of \emph{attributes} and  their \emph{values} and 
collect multiple corresponding natural language texts (references) -- utterances consisting of one or several sentences. An example MR-reference pair is shown in Figure~\ref{fig:pair}, Table~\ref{tab:attr} lists all the attributes in our domain.

\begin{figure}[t]
\begin{center}
\small
\setlength{\extrarowheight}{3pt}
\begin{tabular}{c>{\raggedright\hangindent=2em\arraybackslash}m{5cm}}
\bf MR & name[The Wrestlers], priceRange[cheap], customerRating[low] \\\hline
\bf Reference & The wrestlers offers competitive prices, but isn't highly rated by customers.\\
\end{tabular}
\end{center}
\caption{Example of an MR-reference pair.} 
\label{fig:pair}
\end{figure}


\begin{table}[tb]
\begin{center}
\small
\setlength{\extrarowheight}{2pt}
\begin{tabular}{>{\hspace{-1mm}}l>{\hspace{-2mm}}l>{\hspace{-2mm}}l}
\textbf{Attribute} & \textbf{Data Type} & \textbf{Example value}\\ \hline \hline
name & \hspace{-3mm}verbatim string & \it The Eagle, ...\\  
eatType & dictionary & \it restaurant, pub, ...\\  
familyFriendly & boolean & \it Yes / No\\  
priceRange & dictionary & \it cheap, expensive, ...\\  
food & dictionary & \it French, Italian, ...\\  
near & \hspace{-3mm}verbatim string & \it Zizzi, Cafe Adriatic, ...\\
area & dictionary & \it riverside, city center, ...\\  
customerRating & dictionary & \it 1\,of\,5\,(low),\,4\,of\,5\,(high),\,...\hspace{-5mm} \\  
\end{tabular}
\end{center}
\caption{Domain ontology of the E2E dataset.}
\label{tab:attr}
\end{table}%

\begin{figure}[t]
\begin{center}
\includegraphics[height=2.7cm]{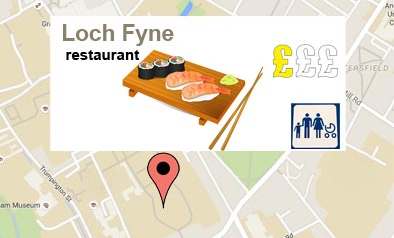}
\end{center}
\caption{An example pictorial MR.}
\label{fig:picture}
\end{figure}

In contrast to previous work \cite{mairesse:acl2010,wen_stochastic_2015,dusek_context-aware_2016}, we use different modalities of meaning representation for data collection: textual/logical and pictorial MRs. 
The textual/logical MRs (see Figure~\ref{fig:pair}) take the form of a sequence with attribute-value pairs provided in a random order. 
The pictorial MRs (see Figure \ref{fig:picture}) are semi-automatically generated pictures with a combination of icons corresponding to the appropriate attributes. The icons are located on a background showing a map of a city, thus allowing to represent the meaning of attributes \emph{area} and \emph{near} (cf.~Table~\ref{tab:attr}).

In a pre-study \cite{novikova:INLG2016}, we showed that pictorial MRs provide similar collection speed and utterance length, but are less likely to prime the crowd workers in their lexical choices. Utterances produced using pictorial MRs were considered to be more informative, natural and better phrased.
However, while pictorial MRs provide more variety in the utterances, this also introduces noise.
Therefore, we decided to use pictorial MRs to collect 20\% of the dataset.

Our crowd workers  were asked to verbalise all 
information from the MR; however, they were not penalised for skipping an attribute.
This makes the dataset more challenging, as NLG systems 
need to account for noise in training data. 
On the other hand, the systems are helped by having multiple human references per MR at their disposal.

\subsection{Data Statistics} 

The resulting dataset \cite{novikova_e2e_2017} contains over 50k references for 6k distinct MRs (cf.\ Table~\ref{tab:dataset-stats}), which is 10 times bigger than previous sets in comparable domains (BAGEL, SF Hotels/Restaurants, RoboCup). The dataset contains more human references per MR (8.27 on average), which should make it more suitable for data-driven approaches. 
However, it is also more challenging as it uses a larger number of sentences in references (up to 6 compared to 1--2 in other sets) and more attributes in MRs.

\begin{table}[tb]
\centering
\small
\setlength{\extrarowheight}{2pt}
\begin{tabular}{lcc}
\textbf{E2E data part} & \textbf{MRs} & \textbf{References} \\ \hline \hline
training set & 4,862 & 42,061  \\
development set & \phantom{0,}547 & \phantom{0}4,672 \\
test set & \phantom{0,}630 & \phantom{0}4,693 \\\hdashline[0.5pt/2pt]
full dataset & 6,039 & 51,426
\end{tabular}
\caption{Total number of MRs and human references in the E2E data sections.}
\label{tab:dataset-stats}
\end{table}

For the E2E challenge, we split the data into training, development and test sets (in a roughly 82-9-9 ratio). MRs in the test set are all previously unseen, i.e.\ none of them overlaps with training/development sets, even if restaurant names are removed. MRs for the test set were only released to participants two weeks before the challenge submission deadline on October 31, 2017. Participants had no access to test reference texts.
The whole dataset is now freely available at the E2E NLG Challenge website 
at:
\begin{center}
\url{http://www.macs.hw.ac.uk/InteractionLab/E2E/}
\end{center}

\section{Systems in the Competition}\label{sec:systems}

\begin{table*}[t]
\begin{center}
\scriptsize
\setlength{\extrarowheight}{3pt}
\begin{tabular}{>{\hspace{-1mm}\raggedright\hangindent=2em\arraybackslash}m{8.8cm}>{\hspace{-1mm}}cccccccc}
\textbf{System} & \bf BLEU & \bf NIST & \bf \hspace{-2mm}METEOR\hspace{-1mm} & \bf \hspace{-2mm}ROUGE-L\hspace{-3mm} & \bf CIDEr & \bf \hspace{-2mm}norm.~avg.\hspace{-2mm} \\ \hline\hline
\Ctgen\ baseline~\cite{novikova_e2e_2017}: seq2seq with MR classifier reranking & 0.6593  & 8.6094  & 0.4483  & 0.6850  & 2.2338 & 0.5754 \\\hdashline[0.5pt/2pt]
\Cslug~\cite{juraska_slug2slug:_2018}: seq2seq-based ensemble (LSTM/CNN encoders, LSTM decoder), heuristic slot aligner reranking, data augmentation & \bf 0.6619  & \bf 8.6130  & 0.4454  & 0.6772  & \bf 2.2615 & 0.5744 \\
\Ctntnlgi~\cite{oraby_tntnlg-personage_2018}: \tgen with data augmentation  & 0.6561  & 8.5105  & \bf 0.4517  & 0.6839  & 2.2183 & 0.5729 \\
\Cnle~\cite{agarwal_char-based_2018}: fully lexicalised character-based seq2seq with MR classification reranking & 0.6534  & 8.5300  & 0.4435  & 0.6829  & 2.1539 & 0.5696 \\
\Ctntnlgii~\cite{tandon_tntnlg-mr_shuffle_2018}: \tgen with data augmentation & 0.6502  & 8.5211  & 0.4396  & \bf 0.6853  & 2.1670 & 0.5688 \\
\Charv~\cite{gehrmann_end--end_2018}: fully lexicalised seq2seq with copy mechanism, coverage penalty reranking, diverse ensembling  & 0.6496  & 8.5268  & 0.4386  & \bf 0.6872  & 2.0850 & 0.5673 \\
\Czhang~\cite{zhang_attention_2018}: fully lexicalised seq2seq over subword units, attention memory & 0.6545  & 8.1840  & 0.4392  &\bf 0.7083  & 2.1012 & 0.5661 \\
\Cgong~\cite{gong_technical_2018}: \tgen fine-tuned using reinforcement learning  & 0.6422  & 8.3453  & 0.4469  & 0.6645  & \bf 2.2721 & 0.5631 \\
\Cthomsoni~\cite{schilder_e2e_2018}: seq2seq with stronger delexicalization (incl.\ \emph{priceRange} and \emph{customerRating})  & 0.6336  & 8.1848  & 0.4322  & 0.6828  & 2.1425 & 0.5563 \\
\Csheffi~\cite{chen_shefeld_2018}: 2-level linear classifiers deciding on next slot/token, trained using LOLS, training data filtering & 0.6015  & 8.3075  & 0.4405  & 0.6778  & 2.1775 & 0.5537 \\
\Cdangnt~\cite{nguyen_structure-based_2018}: rule-based two-step approach, selecting phrases for each slot + lexicalising  & 0.5990  & 7.9277  & 0.4346  & 0.6634  & 2.0783 & 0.5395 \\
\Cslugalt~\cite[\emph{late submission,}][]{juraska_slug2slug:_2018}: \slug trained only using complex sentences from the training data & 0.6035  & 8.3954  & 0.4369  & 0.5991  & 2.1019 & 0.5378 \\
\Czhawii~\cite{deriu_end--end_2018}: semantically conditioned LSTM RNN language model \cite{wen:emnlp2015} + controlling the first generated word  & 0.6004  & 8.1394  & 0.4388  & 0.6119  & 1.9188 & 0.5314 \\
\Ctuda~\cite{puzikov_e2e_2018}: handcrafted templates   & 0.5657  & 7.4544  & \bf 0.4529  & 0.6614  & 1.8206 & 0.5215 \\
\Czhawi~\cite{deriu_end--end_2018}: \zhawii with MR classification loss + reranking  & 0.5864  & 8.0212  & 0.4322  & 0.5998  & 1.8173 & 0.5205 \\
\Cadapt~\cite{elder_e2e_2018}: seq2seq with preprocessing that enriches the MR with desired target words & 0.5092  & 7.1954  & 0.4025  & 0.5872  & 1.5039 & 0.4738 \\
\Cchen~\cite{chen_general_2018}: fully lexicalised seq2seq with copy mechanism and attention memory  & 0.5859  & 5.4383  & 0.3836  & 0.6714  & 1.5790 & 0.4685 \\
\Cforgeiii~\cite{mille_forge_2018}: templates mined from training data   & 0.4599  & 7.1092  & 0.3858  & 0.5611  & 1.5586 & 0.4547 \\
\Csheffii~\cite{chen_shefeld_2018}: vanilla seq2seq  & 0.5436  & 5.7462  & 0.3561  & 0.6152  & 1.4130 & 0.4462 \\
\Cthomsonii~\cite{schilder_e2e_2018}: templates mined from training data   & 0.4202  & 6.7686  & 0.3968  & 0.5481  & 1.4389 & 0.4372 \\
\Cforgei~\cite{mille_forge_2018}: grammar-based  & 0.4207  & 6.5139  & 0.3685  & 0.5437  & 1.3106 & 0.4231 \\
\end{tabular}%
\end{center}
\caption{A list of primary systems in the E2E NLG challenge, with word-overlap metric scores.}
\label{tab:primary-systems-wbms}

\medskip\small
System architectures are coded with colours and symbols: \textcolor{seqtoseq}{\symbseq seq2seq}, \textcolor{datadriven}{\symbdd other data-driven}, \textcolor{rules}{\symbrule rule-based}, \textcolor{templates}{\symbtempl template-based}.
Unless noted otherwise, all data-driven systems use partial delexicalisation (with \emph{name} and \emph{near} attributes replaced by placeholders during generation), template- and rule-based systems delexicalise all attributes.
In addition to word-overlap metrics (see Section~\ref{sec:results-automatic}), we show the average of all metrics' values normalised into the 0-1 range, and use this to sort the list. Any values higher than the baseline are marked in bold.
\end{table*}

The interest in the E2E Challenge has by far exceeded our expectations. We received a total of 62 submitted systems by 17 institutions (about 1/3 from industry).
In accordance with ethical considerations for NLP shared tasks \cite{ethicalSharedTasks}, we allowed researchers to withdraw or anonymise their results if their system performs in the lower 50\% of submissions.
Two groups from industry withdrew their submissions and one group asked to be anonymised
after obtaining automatic evaluation results. 

We asked each of the remaining teams to identify 1-2 primary systems, which resulted in 20 systems by 14 groups. 
Each primary system is described in a short technical paper (available on the challenge website) and was
 evaluated both by automatic metrics and human judges (see Section~\ref{sec:results}).
We compare the primary systems to a baseline based on the \tgen generator \cite{Dusek:ACL16}.
An overview of all primary systems is given in Table~\ref{tab:primary-systems-wbms}, including the main features of their architectures.
A more detailed description and comparison of systems will be given in \cite{dusek2018e2e}.

\section{Evaluation Results}\label{sec:results}

\begin{table*}[tp]
\begin{center}
\scriptsize
\setlength{\extrarowheight}{2pt}
\rotatebox[origin=c]{90}{\bf\small Quality}
\
\begin{tabular}{ccr@{--}ll}
\bf \#  & \bf TrueSkill  & \multicolumn{2}{c}{\bf Rank}  & \bf System \\\hline\hline
1  & \phantom{-}0.300  & 1 & 1  & \Cslug \\\hdashline[0.5pt/2pt]
\multirow{13}{*}{2}  & \phantom{-}0.228  & 2 & 4  & \Ctuda \\
& \phantom{-}0.213  & 2 & 5  & \Cgong \\
& \phantom{-}0.184  & 3 & 5  & \Cdangnt \\
& \phantom{-}0.184  & 3 & 6  & \Ctgen \\
& \phantom{-}0.136  & 5 & 7  & \Cslugalt~\emph{(late)} \\
& \phantom{-}0.117  & 6 & 8  & \Czhawii \\
& \phantom{-}0.084  & 7 & 10  & \Ctntnlgi \\
& \phantom{-}0.065  & 8 & 10  & \Ctntnlgii \\
& \phantom{-}0.048  & 8 & 12  & \Cnle \\
& \phantom{-}0.018  & 10 & 13  & \Czhawi \\
& \phantom{-}0.014  & 10 & 14  & \Cforgei \\
& -0.012  & 11 & 14  & \Csheffi \\
& -0.012  & 11 & 14  & \Charv \\\hdashline[0.5pt/2pt]
\multirow{2}{*}{3}  & -0.078  & 15 & 16  & \Cthomsonii \\
& -0.083  & 15 & 16  & \Cforgeiii \\\hdashline[0.5pt/2pt]
\multirow{3}{*}{4}  & -0.152  & 17 & 19  & \Cadapt \\
& -0.185  & 17 & 19  & \Cthomsoni \\
& -0.186  & 17 & 19  & \Czhang \\\hdashline[0.5pt/2pt]
\multirow{2}{*}{5}  & -0.426  & 20 & 21  & \Cchen \\
& -0.457  & 20 & 21  & \Csheffii \\
\end{tabular}
\qquad\qquad
\rotatebox[origin=c]{90}{\bf\small Naturalness}
\  
\begin{tabular}{ccr@{--}ll}
\bf \#  & \bf TrueSkill  & \multicolumn{2}{c}{\bf Rank}  & \bf System \\\hline\hline
1  & \phantom{-}0.211  & 1 & 1  & \Csheffii \\\hdashline[0.5pt/2pt]
\multirow{11}{*}{2}  & \phantom{-}0.171  & 2 & 3  & \Cslug \\
& \phantom{-}0.154  & 2 & 4  & \Cchen \\
& \phantom{-}0.126  & 3 & 6  & \Charv \\
& \phantom{-}0.105  & 4 & 8  & \Cnle \\
& \phantom{-}0.101  & 4 & 8  & \Ctgen \\
& \phantom{-}0.091  & 5 & 8  & \Cdangnt \\
& \phantom{-}0.077  & 5 & 10  & \Ctuda \\
& \phantom{-}0.060  & 7 & 11  & \Ctntnlgii \\
& \phantom{-}0.046  & 9 & 12  & \Cgong \\
& \phantom{-}0.027  & 9 & 12  & \Ctntnlgi \\
& \phantom{-}0.027  & 10 & 12  & \Czhang \\\hdashline[0.5pt/2pt]
\multirow{5}{*}{3}  & -0.053  & 13 & 16  & \Cthomsoni \\
& -0.073  & 13 & 17  & \Cslugalt~\emph{(late)} \\
& -0.077  & 13 & 17  & \Csheffi \\
& -0.083  & 13 & 17  & \Czhawii \\
& -0.104  & 15 & 17  & \Czhawi \\\hdashline[0.5pt/2pt]
\multirow{2}{*}{4}  & -0.144  & 18 & 19  & \Cforgei \\
& -0.164  & 18 & 19  & \Cadapt \\\hdashline[0.5pt/2pt]
\multirow{2}{*}{5}  & -0.243  & 20 & 21  & \Cthomsonii \\
& -0.255  & 20 & 21  & \Cforgeiii \\
\end{tabular}
\end{center}
\caption{TrueSkill measurements of \emph{quality} (left) and \emph{naturalness} (right).}\label{tab:trueskill}

\small
Significance cluster number, TrueSkill value, range of ranks where the system falls in 95\% of cases or more, system name. Significance clusters are separated by a dotted line. Systems are colour-coded by architecture as in Table~\ref{tab:primary-systems-wbms}.
\end{table*}

\subsection{Word-overlap Metrics}
\label{sec:results-automatic}

Following previous shared tasks in related fields \cite{bojar_results_2017,chen_microsoft_2015}, we selected a range of 
metrics measuring word-overlap between system output and references, including BLEU, NIST, METEOR, ROUGE-L, and CIDEr.
%
Table~\ref{tab:primary-systems-wbms} summarises the primary system scores. The \tgen baseline is very strong in terms of word-overlap metrics: No primary system is able to beat it in terms of all metrics 
-- only \slug comes very close. Several other systems beat \tgen in one of the metrics but not in others.\footnote{%
Note, however, that several secondary system submissions perform better than the primary ones (and the baseline) with respect to word-overlap metrics. 
}
Overall, seq2seq-based systems show the best word-based metric values, followed by \sheffi, a data-driven system based on imitation learning. 
Template-based and rule-based systems mostly score at the bottom of the list.  

\subsection{Results of Human Evaluation}
\label{sec:results-human}

However, the human evaluation study provides a different picture.
Rank-based Magnitude Estimation (RankME)~\cite{novikova:2018} was used for evaluation, where crowd workers compared outputs of 5 systems for the same MR and assigned scores on a continuous scale.
We evaluated output naturalness and overall quality in separate tasks; for naturalness evaluation, the source MR was not shown to workers.
We collected 4,239 5-way rankings for naturalness and 2,979 for quality, comparing 9.5 systems per MR on average.

The final evaluation results were produced using the TrueSkill algorithm ~\cite{herbrich2007trueskill, sakaguchi2014efficient}, with partial ordering into significance clusters computed using bootstrap resampling \cite{bojar_findings_2013,bojar_findings_2014,sakaguchi2014efficient}. 
For both criteria, this resulted in 5 clusters of systems with significantly different performance and showed 
a clear winner: \sheffii for {\em naturalness} and \slug for {\em quality}. 
The 2nd clusters 
are quite large for both criteria 
-- they contain 13 and 11 systems, respectively, and both include the baseline \tgen system.

The results indicate that seq2seq systems dominate in terms of {\em naturalness} of their outputs, while most systems of other architectures score lower. The bottom cluster is filled with template-based systems. 
The results for {\em quality} are, however, more mixed in terms of architectures, with none of them clearly prevailing. 
Here, seq2seq systems with reranking based on checking output correctness score high while seq2seq systems with no such mechanism occupy the bottom two clusters.

\section{Conclusion}\label{sec:conclusion}

This paper presents the first shared task on end-to-end NLG. 
The aim of this challenge was to assess the capabilities of recent end-to-end, fully data-driven NLG systems, which can be trained from pairs of input MRs and texts, without the need for fine-grained semantic alignments.
We created a novel dataset for the challenge, which is an order-of-magnitude bigger than any previous publicly available dataset for task-oriented NLG. 
We received 62 system submissions by 17 participating institutions,
with a wide range of architectures, from seq2seq-based models to simple templates.
We evaluated all the entries in terms of five different automatic metrics; 20 primary submissions (as identified by the 14 remaining participants) underwent crowdsourced human evaluation of naturalness and overall quality of their outputs.
 
We consider the \slug system \cite{juraska_slug2slug:_2018}, a seq2seq-based ensemble system with a reranker, as the overall winner of the E2E NLG challenge.
\slug scores best in human evaluations of quality, it is placed in the 2nd-best cluster of systems in terms of naturalness and reaches high automatic scores. 
While the \sheffii system \cite{chen_shefeld_2018}, a vanilla seq2seq setup, won in terms of naturalness, it scores poorly on overall quality -- it placed in the last cluster. 
The \tgen baseline system turned out hard to beat: It ranked highest on average in word-overlap-based automatic metrics and placed in the 2nd cluster in both quality and naturalness.

The results in general show the seq2seq architecture as very capable, 
but requiring reranking to reach high-quality results.
On the other hand, while rule-based approaches are not able to beat data-driven systems in terms of automatic metrics, they often perform comparably or better in human evaluations.


We are preparing a detailed analysis of the results \cite{dusek2018e2e} and a release of all system outputs with user ratings on the challenge website.\footnote{\url{http://www.macs.hw.ac.uk/InteractionLab/E2E}} We plan to use this data for experiments in automatic NLG output quality estimation \cite{specia:MT2010,dusek_referenceless_2017}, where the large amount of data obtained in this challenge allows a wider range of experiments than previously possible.

\subsection*{Acknowledgements}

This research received funding from the EPSRC projects  DILiGENt (EP/M005429/1) and  MaDrIgAL (EP/N017536/1). The Titan Xp used for this research was donated by the NVIDIA Corporation.

\bibliography{inlg2018}
\bibliographystyle{acl_natbib}

\end{document}